\documentclass{article}

\usepackage{PRIMEarxiv}

\usepackage[utf8]{inputenc} 
\usepackage[T1]{fontenc}    
\usepackage{hyperref}       
\usepackage{url}            
\usepackage{booktabs}       
\usepackage{amsfonts}       
\usepackage{nicefrac}       
\usepackage{microtype}      
\usepackage{lipsum}
\usepackage{fancyhdr}       
\usepackage{graphicx}       
\graphicspath{{media/}}     
\usepackage{tabularx}
\usepackage{booktabs} 
\usepackage{array}    
\usepackage{pifont}  
\usepackage{pdflscape} 
\usepackage{tikz}
\usetikzlibrary{arrows.meta, positioning, shapes.multipart, calc}
\usetikzlibrary{decorations.pathreplacing}
\usepackage{lmodern} 
\usepackage{float}

\title{Recalibrating the Compass: Integrating Large Language Models into Classical Research Methods}

\author{
    \begin{minipage}[t]{0.5\textwidth}
        \centering
        Tai-Quan ~Peng\thanks{Corresponding Author: \texttt{winsonpeng@gmail.com}} \\
        Department of Communication\\
        Michigan State University\\
        East Lansing, MI 48824 
    \end{minipage}
    \hfill
    \begin{minipage}[t]{0.5\textwidth}
        \centering
        Xuzhen ~Yang \\
        Department of Communication\\
        Michigan State University\\
        East Lansing, MI 48824 
    \end{minipage}
}

\begin{document}
\maketitle
\vspace{-1em}
\begin{abstract}
This paper examines how large language models (LLMs) are transforming core quantitative methods in communication research in particular, and in the social sciences more broadly—namely, content analysis, survey research, and experimental studies. Rather than replacing classical approaches, LLMs introduce new possibilities for coding and interpreting text, simulating dynamic respondents, and generating personalized and interactive stimuli. Drawing on recent interdisciplinary work, the paper highlights both the potential and limitations of LLMs as research tools, including issues of validity, bias, and interpretability. To situate these developments theoretically, the paper revisits Lasswell’s foundational framework—“Who says what, in which channel, to whom, with what effect?”—and demonstrates how LLMs reconfigure message studies, audience analysis, and effects research by enabling interpretive variation, audience trajectory modeling, and counterfactual experimentation. Revisiting the metaphor of the methodological compass, the paper argues that classical research logics remain essential as the field integrates LLMs and generative AI. By treating LLMs not only as technical instruments but also as epistemic and cultural tools, the paper calls for thoughtful, rigorous, and imaginative use of LLMs in future communication and social science research.
\end{abstract}

\keywords{Large Language Models \and Survey \and Experiment \and Content Analysis\and Computational Social Science}

\vspace{0.5em}
\makeatletter
\begin{center}
\today
\end{center}
\makeatother

\setcounter{tocdepth}{1}
\tableofcontents
\newpage 

\section{Introduction}
Social science research—especially in communication research—has long relied on three core methods: surveys, experiments, and content analysis. These tools have shaped how scholars study attitudes, behaviors, causal mechanisms, and media discourse. While recent technologies have brought new types of data and analysis, traditional methods still anchor how researchers ask questions and make sense of the world.

\begin{figure}[h!]
\centering
\begin{tikzpicture}[
    box/.style={rectangle, rounded corners, draw=black, text width=2.8cm, align=left, font=\scriptsize, fill=#1!20, minimum height=2.8cm},
    dot/.style={circle, draw=black, fill=red, inner sep=1.5pt},
    label/.style={font=\footnotesize},
    ->, thick
]

\def\xA{1.3}
\def\xB{4.8}
\def\xC{8.3}
\def\xD{11.8}
\def\xE{15.3}

\draw[->, thick] (\xA - 1, 0) -- (16.5,0) node[right] {Future};

\node[circle, draw=black, fill=pink!30, inner sep=1.5pt] at (\xA, 0) {};
\node[label] at (\xA, -0.5) {Pre-Internet};

\node[circle, draw=black, fill=orange!30, inner sep=1.5pt] at (\xB, 0) {};
\node[label] at (\xB, -0.5) {1990s};

\node[circle, draw=black, fill=yellow!30, inner sep=1.5pt] at (\xC, 0) {};
\node[label] at (\xC, -0.5) {2000s};

\node[circle, draw=black, fill=lime!30, inner sep=1.5pt] at (\xD, 0) {};
\node[label] at (\xD, -0.5) {2010s};

\node[circle, draw=black, fill=cyan!30, inner sep=1.5pt] at (\xE, 0) {};
\node[label] at (\xE, -0.5) {2020s};

\node[box=pink] at (\xA, 2) {\textbf{Pre–Internet}\\
1. Linear consumption of media information\\
2. Minimal interaction or feedback mechanism};

\node[box=orange] at (\xB, 2) {\textbf{Internet}\\
1. Hyperlink\\
2. Emergence of interactivity, but very preliminary\\
3. From consumption to engagement};

\node[box=yellow] at (\xC, 2) {\textbf{Social Media}\\
1. Enhanced participation and content creation\\
2. Empowered audience to connect, share, and influence on a larger scale};

\node[box=lime] at (\xD, 2) {\textbf{Mobile Media}\\
1. On-the-go interaction \\
2. Ubiquitous access to media \& personalized content\\
3. Enhanced audience control and immediacy};

\node[box=cyan] at (\xE, 2) {\textbf{Generative AI}\\
1. Advanced personalization\\
2. Predictive content delivery\\
3. Interactive AI-driven experiences};

\node[box=pink] at (\xA, -2.5) {\textbf{Quantitative Methods}\\
Surveys\\
Experiments\\
Content analysis};

\node[box=orange] at (\xB, -2.5) {\textbf{Web Analytics}\\
Clickstream Analysis\\
Hyperlink Analysis\\
Log analysis};

\node[box=yellow] at (\xC, -2.5) {\textbf{Social Media Analytics}\\
Text Mining\\
Network Analysis\\
Online Experiment};

\node[box=lime] at (\xD, -2.5) {\textbf{Mobile Analytics}\\
Mobile Sensing\\
Spatial Modeling};

\node[box=cyan] at (\xE, -2.5) {\textbf{Emerging Methods}\\
LLM-powered Content Analysis\\
LLM-augmented Survey\\
LLM-catalyzed Experiment\\
LLM-enabled Simulation};

\draw [decorate,decoration={brace,mirror,amplitude=6pt},thick, -]
  (\xB - 1.4, -4.1) -- (\xD + 1.4, -4.1)
  node[midway, below=6pt, font=\footnotesize] {Computational Social Science (CSS)};
  
\draw [decorate,decoration={brace,mirror,amplitude=6pt},thick, -]
  (\xE - 1.4, -4.1) -- (\xE + 1.4, -4.1)
  node[midway, below=6pt, font=\footnotesize] {AI-Enhanced CSS};

\end{tikzpicture}
\caption{Co-evolution of Media Technologies and Research Methods}
\label{fig:evolution}
\end{figure}
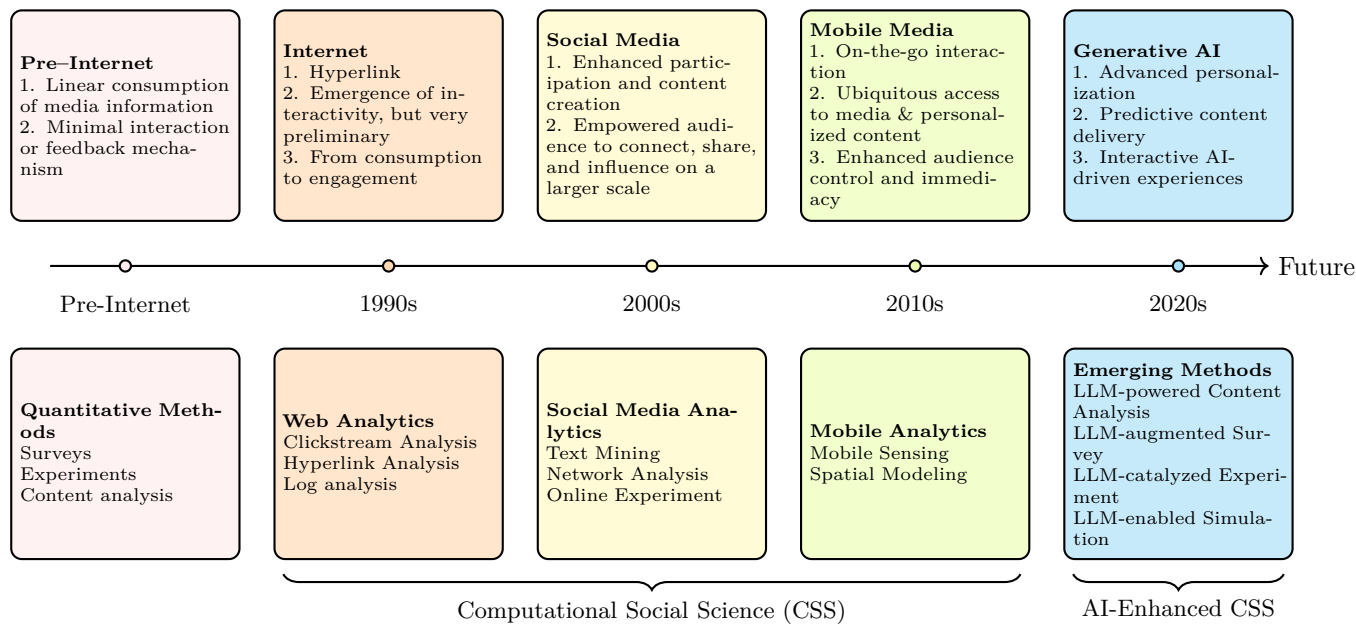

The past decades have witnessed significant shifts in media technologies. As Figure \ref{fig:evolution} shows, social science research has evolved in tandem with these changes. From the early days of mass media to the rise of the web, social platforms, mobile devices, and now generative artificial intelligence (AI), communication has become more personalized, interactive, and predictive. Compared to scholars in other fields, communication scholars have been especially responsive to these changes, often at the forefront of adapting research methods to evolving media environments. Methodological approaches have expanded accordingly—from classical research designs through web, social media, and mobile analytics, to the most recent techniques powered by large language models (LLMs). 

Rather than making earlier tools obsolete, these changes add new layers to established practices. Researchers today work in a more complex methodological environment, but the value of interpretability, theoretical grounding, and careful design remains central. One way to think about this is as a kind of \textit{methodological compass}—a guide for navigating the expanding array of methodological options, alongside the uncertainties and unknowns that accompany them. Surveys help make sense of behavioral traces, experiments put algorithmic predictions to the test, and manual coding supports the interpretation of automated outputs, including those from LLMs. In this landscape, classical and computational methods work together, often reinforcing each other’s strengths.

LLMs are a major development in this ongoing evolution. These LLM-enabled approaches, including content analysis, survey augmentation, and simulation-based experiments, don’t replace existing methods. Their flexibility opens up new possibilities: generating experimental stimuli, simulating survey responses, interpreting open-ended answers, and supporting dialogic interaction. When used carefully, they can extend familiar methods in creative ways and expand what researchers can observe, test, and interpret. But their use also raises important questions—about what counts as valid inference, how much control researchers have, and what it means to do empirical work with machines that “speak.”

LLMs' influence on social science research transcends disciplinary boundaries. Their implications are not confined to communication research alone. Across fields like political science, psychology, public health, and computer science, LLMs are being used to create synthetic populations, simulate interactions, and assist with qualitative analysis. As Lazer and colleagues \cite{lazer_computational_2009} pointed out more than a decade ago, the real challenge isn’t just building new tools—it’s integrating them into research in a way that supports cumulative knowledge. That insight still holds today. As van Atteveldt and Peng \cite{van_atteveldt_when_2018} argued, computational approaches work best when they align with the logics of established research traditions.

This paper builds on that foundation. While it draws from communication, it takes an interdisciplinary view, considering how LLMs are reshaping empirical research more broadly. By mapping current practices and conceptual shifts, the paper joins ongoing conversations about research design, methodological rigor, and the evolving role of classical methods in an AI-driven era.

The discussion focuses on three areas where LLMs are already making an impact: content analysis, surveys, and experiments. The first section looks at how LLMs serve not only as automated coders but also as tools for framing and interpretation. The second explores their role in simulating respondents, estimating opinions across groups, and testing question design. The third turns to experimental research, including personalized message generation, real-time interaction, and counterfactual scenarios. A concluding section revisits Lasswell’s “5Ws” to consider how LLMs are reshaping how we study messages, audiences, and effects. Across these domains, we argue that LLMs offer new affordances while underscoring the continued relevance of classical approaches as anchors for empirical inquiry.

\section{LLMs as Language Tools for Content Analysis}
\label{sec:headings}

LLMs are transforming content analysis by functioning as coders, classifiers, and interpreters of textual data. Traditionally, content analysis has relied on manual coding or rule-based algorithms such as dictionaries or supervised classifiers. These methods, while effective, face limitations in scalability, adaptability, and sensitivity to linguistic nuance. LLMs—particularly models like ChatGPT—offer a compelling alternative: zero-shot or few-shot classification using natural language prompts. This capability enables researchers to extract complex features from text (e.g., topics, frames, sentiment) without the need for extensive labeled data or retraining, offering both theoretical flexibility and computational efficiency.

Recent empirical studies highlight the versatility of LLMs across a broad range of annotation tasks. For example, Gilardi et al. \cite{gilardi_chatgpt_2023} demonstrate that ChatGPT outperforms both crowd workers and trained annotators in relevance, stance, topic, and frame detection across four datasets of tweets and news articles. In a related application, Li et al. \cite{li_hot_2024} show that ChatGPT achieves roughly 80\% accuracy in detecting hateful, offensive, and toxic (HOT) content relative to MTurker ground truth—though it is notably more consistent in identifying non-HOT content than HOT content. In open-ended survey coding, Mellon et al. \cite{mellon_ais_2023} find that LLMs approach human-level accuracy when asked to identify “most important issue” responses, providing evidence for their viability in structured classification tasks. However, in normatively charged domains such as fact-checking, performance can falter. DeVerna et al. \cite{deverna_fact-checking_2024} demonstrate that LLM-generated fact checks may actually impair headline discernment, highlighting the risks of overconfidence in AI-mediated annotation.

Beyond their use as efficient annotators, LLMs offer a novel lens for reimagining content analysis as a site of interpretive variation rather than consensus. In our recent study \cite{kang_embracing_2025}, we demonstrate how LLMs can be used to simulate multiple audience perspectives—political, demographic, or ideological—when coding complex political texts. Rather than seeking convergence among coders, our approach embraces dialectic intersubjectivity, a process that foregrounds the meaningful divergence in how different social actors interpret the same message. By prompting the model to represent distinct viewpoints (e.g., liberal vs. conservative), we were able to trace how content acquires layered meanings across contexts. This perspective aligns with findings by Gielens et al. \cite{gielens_goodbye_2025}, who show that even when using clearly defined annotation schemes, ChatGPT’s performance varies significantly across argument types—suggesting that such variation may reveal, rather than obscure, the contested nature of public discourse. This application shifts the methodological goal of content analysis from achieving intercoder reliability to understanding and coordinating divergent readings. In this light, LLMs are not simply tools for annotation—they are instruments for surfacing the pluralism of meaning at the heart of human communication.

\subsection{When LLMs Work—and When They Don’t?}
LLMs tend to excel at structured, low-context annotation tasks where categories are discrete and textual features align with well-defined labels—such as detecting stance, relevance, or issue topics. In these domains, models like ChatGPT exhibit strong agreement with human-coded benchmarks and lower performance variance across samples \cite{gilardi_chatgpt_2023}. However, LLMs are less reliable when tasked with identifying subjective, contested, or emotionally charged content. For instance, while the model performs well in identifying neutral or mildly offensive content, its consistency declines when coding highly hateful or ambiguous expressions \cite{li_hot_2024}. Similarly, LLMs used for fact-checking generate coherent and plausible-sounding explanations but may misclassify true statements or introduce epistemic ambiguity \cite{deverna_fact-checking_2024}.

These patterns suggest that LLMs perform best when semantic structure is strong and moral ambiguity is low. Tasks that involve value judgments, cultural interpretation, or truth claims beyond the training distribution challenge the model's generalization capabilities. The implication is that researchers should be cautious about overextending LLMs to normative tasks without robust validation. This pattern of variability also aligns with findings by Calderon et al. \cite{calderon_alternative_2025}, who propose the Alternative Annotator Test (Alt-Test) as a statistical framework for evaluating LLMs as annotators. Rather than expecting LLMs to replicate individual human labels, they argue for comparing LLM outputs to the distribution of human annotations. This approach recognizes that even among trained annotators, disagreement can be meaningful, and LLMs may contribute by offering a reliable approximation of pluralistic human judgment—especially in cases where interpretation is contested or multidimensional.

\subsection{What Affects LLM Annotation Performance?}
Multiple factors influence how LLMs perform as coders in content analysis, as summarized in Table \ref{tab:llm_annotation}. First, model architecture and size matter. Larger models typically produce more fluent and accurate annotations but are not universally superior—especially when prompts are poorly designed or tasks are highly contextual. Second, prompt complexity and robustness plays a key role. Mu et al. \cite{mu_navigating_2023} find that simpler prompts tend to yield more reliable results than multi-step or overly verbose instructions. Huang et al. \cite{huang_large_2023} and Lu et al. \cite{lu_fantastically_2022} further show that complex reasoning chains or “effective” prompts often fail when transferred across tasks or re-ordered, exposing their fragility. Third, output format influences interpretability and stability. Zhao et al. \cite{zhao_calibrate_2021} show that binary classification and probability estimation lead to different outcomes on the same input, with the latter offering better nuance but lower consistency. 

\begin{table}[ht]
    \centering
    \renewcommand{\arraystretch}{1.2} 
    \begin{tabularx}{\textwidth}{|>{\raggedright\arraybackslash}p{3.5cm}|X|}
        \hline
        \textbf{Factor} & \textbf{Findings / Effects} \\
        \hline
        Task Type & High performance on structured tasks (topic, stance); low on subjective tasks (e.g., hate, truth) \\
        \hline
        Model Size & Larger models perform better overall, but not uniformly; gains depend on task/prompt fit \\
        \hline
        Prompt Complexity & Simpler prompts often outperform longer, multi-step ones \cite{mu_navigating_2023} \\
        \hline
        Prompt Robustness & Prompt effectiveness degrades with re-ordering or re-use across tasks \cite{lu_fantastically_2022, huang_large_2023} \\
        \hline
        In-Context Learning & Order, number, and familiarity of examples affect model outputs \\
        \hline
         Output Format & Binary vs. probabilistic output changes accuracy and interpretability \cite{zhao_calibrate_2021} \\
        \hline
        Domain Sensitivity & LLMs struggle with tasks involving moral, political, or cultural ambiguity \cite{li_culturellm_2024, deverna_fact-checking_2024} \\
        \hline
    \end{tabularx}
    \vspace{0.5em}
    \caption{Summary of LLM Annotation Effectiveness and Influencing Factors}
    \label{tab:llm_annotation}
\end{table}

In-context learning—embedding examples within prompts to guide the model—offers another layer of control but introduces its own vulnerabilities. Zhao et al. \cite{zhao_calibrate_2021} show that even small changes to example ordering can lead to significant shifts in model output distributions—effects analogous to priming in survey design. This fragility poses challenges for replicability and interpretability, especially when prompts are not fully reported or when model behavior shifts across updates. As LLMs become more integrated into communication research, scholars will need to treat prompt design and example construction not merely as technical preprocessing steps, but as core elements of research design. A future methodological agenda should include the development of standardized prompt libraries, reporting guidelines, and diagnostic tools to assess the stability of LLM outputs under different conditions. These results highlight the importance of treating prompt design, model selection, and output format as experimental variables, not background choices. Further complicating the picture is the tradeoff between performance and interpretability. 

Another consideration is the format of the output produced by LLMs, which can significantly affect both the interpretability and reliability of content analysis results. While binary classification (e.g., “Is this hateful? Yes/No”) offers clarity and ease of comparison, it can mask uncertainty and force artificial dichotomies onto nuanced content. In contrast, probability-based formats or ordinal scales allow for gradation and express uncertainty, which may be more faithful to the complexities of language and social meaning. However, these richer formats also introduce analytic challenges—such as increased variability across runs or difficulties in benchmarking against human coders. This tradeoff highlights a central tension in LLM-assisted content analysis: greater flexibility comes at the cost of reduced standardization. Researchers must therefore carefully select output formats that align with the theoretical goals of the analysis, while transparently reporting how LLMs are instructed to produce those outputs.

A last consideration is domain sensitivity—LLMs often struggle with tasks that involve moral, political, or cultural ambiguity. As Li et al. \cite{li_culturellm_2024} show, these models tend to reflect dominant cultural norms, especially those embedded in English-language training data, which can lead to biased interpretations when applied across diverse cultural contexts. DeVerna et al. \cite{deverna_fact-checking_2024} similarly find that in politically charged settings, LLM-generated fact checks can reduce users’ ability to distinguish between true and false headlines, particularly when the model is uncertain or makes incorrect judgments. These findings highlight that LLMs are not culturally neutral tools; their performance can vary depending on the domain, making careful calibration and contextual awareness essential in content analysis.

These dimensions—model choice, prompt construction, example selection, and output formatting—shape not just the reliability of results, but also their transparency. As Feuerriegel et al. \cite{feuerriegel_using_2025} argue, while LLM-based classifiers often outperform traditional dictionary or rule-based tools in predictive accuracy, they do so with significantly reduced transparency. This tradeoff is especially critical in social science research, where researchers must be able to explain how decisions were reached.

These challenges reinforce the importance of documenting prompt construction, output structure, and decision-making logic when using LLMs in scientific workflows. Researchers should document prompt structures, output formats, and example inputs with the same care given to survey instruments or coding protocols. Moving forward, the field would benefit from standardized prompt libraries, reporting guidelines, and diagnostic tools to assess the robustness of LLM outputs under varying conditions. Treating these elements as central to research design—not peripheral—will be essential for the responsible use of LLMs in communication research.

\subsection{Recommended Practices for LLM-powered Content Analysis}
As LLMs become increasingly integrated into content analysis workflows, the central question is not whether they will replace human coding, but how they can best augment it. Rather than treating LLMs as a replacement for trained coders, researchers should adopt a triangulation strategy: using LLM outputs as one layer of interpretation alongside human judgment, traditional coding schemes, or computational classifiers. Such triangulation not only strengthens validity but also creates opportunities to cross-validate results and identify cases of ambiguity, disagreement, or hidden bias. This aligns with broader calls for human-in-the-loop AI \cite{natarajan_human---loop_2024} in social science research, emphasizing the complementary strengths of automated scalability and human contextual sensitivity. Triangulation also provides a practical framework for managing the inherent uncertainty and bias in LLM-generated annotations—especially on tasks involving moral or political content. This approach reinforces the idea that LLMs are not epistemically neutral tools, but interpretation machines whose outputs must be contextualized, audited, and, when needed, challenged.

To advance cumulative knowledge, the field should move toward standardizing prompt design and evaluation practices. Prompts are currently treated as ad hoc instructions—part art, part improvisation—but they need to become formal, shareable components of methodological design. Just as codebooks define categories and decision rules in manual content analysis, future work should develop \textit{LLM-aware codebooks} that include prompt templates, examples, and intended output formats. Such tools will not eliminate the interpretive nature of annotation, but they can make the prompting process more transparent and replicable. 

Comparing outputs across different models—such as GPT, Claude, and open-source systems—can serve as a robustness check. This approach is similar to using intercoder reliability in traditional content analysis. As prompting becomes more systematic, communication scholars can help define how AI-generated meaning is interpreted and used in research. Their involvement is essential for developing clear standards for documentation, validation, and application.

These goals align with emerging standards for evaluating LLMs in social science research. Calderon et al. \cite{calderon_alternative_2025} propose a formal statistical test to determine whether LLMs can approximate the variability found in human coding—a shift from seeking perfect replication to capturing human-like distributions. Meanwhile, Feuerriegel et al. \cite{feuerriegel_using_2025} emphasize that without clear documentation of prompt design and interpretive criteria, the use of LLMs risks undermining the transparency that has historically defined rigorous content analysis. Taken together, these arguments point to the need for more standardized practices—such as shared prompt libraries, reporting protocols, and evaluation tools—designed specifically for communication research.

\section{LLMs as Interview Instruments in Survey Research}

Since 1940s, social scientists have relied heavily on survey methods to measure what people know, believe, and do. Surveys have offered a structured and widely accepted approach to eliciting self-reported attitudes, knowledge, and behaviors from representative samples of the public, forming the empirical backbone of studies in political communication, media effects, and public opinion. However, traditional surveys now face growing challenges. Response rates have steadily declined, threatening representativeness and increasing costs, while the rise of automated bots and low-quality respondents on online survey platforms further compromises data integrity. 

In response, researchers have increasingly turned to computational methods and digital trace data—such as social media posts and search behavior—as alternative ways to observe public expressions and reactions \cite{van_atteveldt_when_2018}. Yet while such data offer scale and immediacy, they often lack the interpretive clarity and respondent control that survey designs provide. Against this backdrop, the emergence of LLMs offers a new alternative: the simulation of diverse human responses to survey questions using generative AI. This technique enables researchers to explore what different types of people might say under controlled conditions—at scale, and with novel flexibility—while also raising new questions about accuracy, bias, and the nature of synthetic data in social science.

\subsection{From Partisanship to Public Life: Broadening the scope of LLM-simulated opinion}
Much of the early research on using LLMs to simulate public opinion concentrated on political contexts, particularly presidential elections, party identification, and ideological divides. This focus was driven in part by the ready availability of benchmark survey data, such as the American National Election Studies (ANES) and Cooperative Election Study (CES), which enabled systematic validation \cite{argyle_out_2023, bisbee_synthetic_2024, kim_ai-augmented_2023}. However, as the field matures, there is growing recognition that LLMs can be used to simulate opinion dynamics in issue domains beyond politics—including climate change, public health, media behavior, and social values \cite{burstein_large_2025, chu_language_2023, lee_can_2024, nguyen_simulating_2024}.

Across this expanding literature, a common thread is that LLMs perform well when the underlying issue is widely discussed online and structured prompts provide sufficient contextual grounding. Studies exploring environmental and health-related topics demonstrate that when LLMs are prompted to simulate individuals with specific demographic profiles or attitudinal orientations, they can produce responses that reflect known public concerns, value conflicts, and even cross-cultural variation \cite{jiang_donald_2025, karanjai_synthesizing_2025, pachot_can_2024}. For instance, in both environmental and global health contexts, LLMs have been shown to generate outputs that match large-scale patterns of vaccine hesitancy or climate concern, particularly when conditioned on ideological or regional cues \cite{burstein_large_2025, lee_can_2024}. These applications underscore the models' potential to serve as low-cost tools for modeling how publics might respond to emerging issues or policy debates, particularly when rapid response is needed and traditional data collection is slow or expensive.

Yet, the shift from political to domain-specific applications also reveals new challenges. One is the risk of over-coherence or stereotype amplification: LLMs may produce plausible but overly homogenized responses that do not adequately capture within-group diversity. This issue becomes especially salient when simulating public opinion on culturally sensitive or morally contested topics such as immigration, abortion, or racial discrimination \cite{qu_performance_2024, salecha_large_2024}. Multiple studies caution that while LLMs can replicate broad ideological patterns, they often fall short in representing nuanced, intersectional, or minority perspectives \cite{abeliuk_fairness_2025, nguyen_simulating_2024}. In education and science communication research, concerns have also been raised about the exaggeration of certain narratives, where personas based on race, gender, or political ideology are rendered too rigid or caricatured \cite{hwang_aligning_2023, nguyen_simulating_2024}. These findings suggest that while LLMs offer scalability and speed, careful prompt design and critical evaluation are necessary to avoid reinforcing oversimplified representations of public opinion.

In sum, the literature is beginning to move beyond binary comparisons of human versus synthetic fidelity, toward deeper reflection on how and when LLMs can validly simulate public reasoning in diverse contexts. The potential is clear: LLMs may eventually support richer, multi-dimensional modeling of attitudes that go beyond vote choice or party identity. But to get there, researchers must continue refining methodological tools, expand the range of validation datasets beyond U.S. politics, and remain attuned to the epistemological and ethical risks of substituting synthetic for lived human opinion \cite{atari_which_2023, pachot_can_2024, santurkar_whose_2023}.

\subsection{From Averages to Identities: Modeling Public Opinion Across Demographic Divides}

Early efforts to validate LLMs as tools for simulating public opinion largely emphasized their ability to replicate aggregate-level response distributions. Studies often evaluated whether synthetic responses could approximate national-level averages on survey items, particularly those related to ideology or partisanship \cite{argyle_out_2023, bisbee_synthetic_2024}. However, this focus on aggregate fidelity obscures a crucial limitation: LLMs often fail to accurately simulate the diversity of opinion within demographic subgroups. As researchers have begun to assess subgroup-level performance, they have consistently found discrepancies across variables such as age, race, income, and education \cite{abeliuk_fairness_2025, qu_performance_2024}.

Several studies show that LLMs simulate more consistent or homogenous responses than those observed in real-world data, particularly among marginalized or intersectional groups. For example, models tend to reflect dominant cultural narratives more reliably than those of underrepresented populations, suggesting that training data imbalances persist in generated outputs \cite{lee_can_2024, santurkar_whose_2023}. Even when models are prompted with detailed demographic attributes, synthetic responses often lack the subtle ideological or experiential variation observed in actual survey data. Research using fine-tuned or persona-conditioned models—such as role-based prompting and retrieval-augmented generation—has improved alignment with human responses in some cases \cite{karanjai_synthesizing_2025}, but group-level prediction gaps remain.

In response, scholars have begun to move from single-variable conditioning to multidimensional approaches that account for interactions between identity variables and attitudes. This shift reflects a broader trend toward demographic realism: simulating opinion heterogeneity within groups, rather than assuming representative averages are sufficient. Still, the field lacks standardized benchmarks for measuring within-group variability in LLM outputs. Until such standards emerge, claims about the demographic generalizability of LLM-simulated opinion must remain provisional.

\subsection{From Fidelity to Bias: Toward a Layered Approach to Bias Diagnosis}
While much early work evaluating LLMs in survey research focused on their ability to replicate aggregate-level human responses, there is growing recognition that fidelity alone is insufficient. Studies have shown that even when LLMs approximate overall response distributions, they may embed structural biases that distort the simulation of public opinion—especially for marginalized groups or sensitive topics \cite{bisbee_synthetic_2024, santurkar_whose_2023, motoki_more_2023}. This growing body of research reflects a shift: from treating fidelity as the gold standard, toward more critical scrutiny of how biases are encoded, compounded, and surfaced in synthetic data.

This perspective is not entirely new. Traditional survey research has long grappled with various forms of bias, including social desirability bias, non-response bias, and coverage error. However, in the context of LLM-simulated data, these biases take on new forms and may arise from sources that are less transparent and more intertwined. To advance methodological rigor, it is essential to disentangle the different layers at which bias can enter the LLM simulation process. Based on recent studies, we identify seven core sources of bias in LLM-based survey research.

\subsubsection{Bias from the Simulated Persona}
Bias can emerge from the way researchers construct and prompt LLMs to adopt specific personas (e.g., "a 30-year-old conservative man from California"). These role definitions may lead models to generate overly coherent or stereotypical responses, especially when prompts rely on coarse identity categories. For example, studies have shown that LLMs often exaggerate ideological consistency or cultural traits when simulating politically or racially marked personas \cite{abeliuk_fairness_2025, nguyen_simulating_2024}. This mirrors concerns in survey methodology where respondent categories—like "Latino voters"—mask within-group heterogeneity.

\subsubsection{Bias from the Model Algorithm}
Even with identical prompts, different LLMs may produce systematically different outputs depending on architectural decisions, model size, and decoding strategies. Shin et al. \cite{shin_ask_2024} demonstrate that various LLMs carry default associations about social groups—e.g., linking political conservatism with specific moral traits—that persist across generations. Santurkar et al. \cite{santurkar_whose_2023} similarly find ideological skew in model responses, with GPT-based models often reflecting centrist or left-leaning assumptions, even under neutral prompts. These algorithmic biases are akin to interviewer effects in survey research: subtle features of the data collection instrument shape the response content.

Building on this line of inquiry, a recent large-scale comparative study offers further insights into the structure and measurement of political bias in LLM-generated responses. Peng et al. \cite{peng_beyond_2025} evaluate 43 LLMs across four geopolitical regions using standardized survey-style prompts to measure model bias. Rather than simulating human personas, the authors adopt a persona-free design to isolate model behavior. Their framework captures both directional bias (e.g., liberal vs. conservative leaning) and differences in sociopolitical engagement—an often overlooked but critical component of public opinion. By benchmarking model responses against real-world human data, the study shows that LLMs not only reflect ideological slant but also reproduce variations in engagement across topics. This work demonstrates that political bias in LLMs is multidimensional and not easily reducible to a single scale, reinforcing the need for nuanced and layered evaluation strategies.

\subsubsection{Bias from the Training Data}
Perhaps the most foundational source of bias lies in the data used to train the LLM. Since most LLMs are trained on internet-scale corpora, they tend to reflect dominant perspectives from English-speaking, educated, and affluent populations. This leads to coverage bias, where underrepresented views—such as those from minority language communities or rural populations—are less accurately simulated \cite{qu_performance_2024, santurkar_whose_2023}. Moreover, widely circulated views online are more likely to shape LLM outputs than those expressed in offline or marginalized spaces, echoing classic concerns about sampling frames in survey design.

Importantly, these training patterns not only affect what the model learns to say, but also how it learns to reason. When exposed to imbalanced or highly correlated input–output associations, LLMs often default to shortcut reasoning—relying on surface-level cues that were predictive in training, even if they lack explanatory power in new contexts. This tendency has been documented across NLP tasks \cite{geirhos_shortcut_2020, mccoy_right_2019} and extends to public opinion modeling. Yang et al. \cite{yang_are_2024} show that LLMs often perform well in predicting survey outcomes (e.g., vote choice) primarily because they exploit tautological features such as party ID or ideological self-placement. Once these shortcut features are removed, performance drops dramatically—revealing that high accuracy often stems from memorized correlations rather than genuine inference. This underscores the importance of distinguishing between surface-level fidelity and substantive validity when using LLMs to simulate survey responses.

\subsubsection{Bias from the Prompt}
Even minor differences in prompt wording, temperature settings, or format can result in measurable changes in LLM output. Several studies have documented how prompt framing effects—such as framing a question as political versus moral—alter simulated responses \cite{pachot_can_2024, tjuatja_llms_2023}. This is similar to question wording effects in traditional surveys, where even neutral phrasing can activate latent attitudes. In LLM-based surveys, these effects may be amplified by the model's sensitivity to initial conditions and instruction tuning.

\subsubsection{Bias from Evaluation Metrics}
Researchers must also consider the role of evaluation metrics in shaping perceptions of fidelity and fairness. Much of the current literature relies on measures such as Pearson correlations, cosine similarity, or regression coefficients to compare LLM outputs to human survey responses \cite{bisbee_synthetic_2024, pachot_can_2024}. While useful, these metrics often prioritize surface-level agreement and may obscure more meaningful discrepancies at the subgroup or semantic level. For example, an LLM may match national-level averages on key items but systematically misrepresent minority or low-frequency views. Moreover, many benchmark datasets used for validation (e.g., CES, GSS, ANES) themselves contain embedded biases due to sampling frames or item wording conventions. These factors risk reinforcing existing assumptions about “correct” opinions, especially when used uncritically to certify LLM performance.

Recent work by Nguyen et al. \cite{nguyen_empirically_2025} pushes this critique further by demonstrating that traditional benchmark metrics can fail to capture the diversity of plausible human responses. In their evaluation of commonsense reasoning, they find that LLM outputs often align with specific annotator subgroups rather than a collective human consensus. As a corrective, they propose comparing model outputs to distributions of human judgments rather than fixed ground-truth labels. This population-based approach underscores the importance of pluralistic benchmarks—especially when evaluating synthetic public opinion data—and cautions against treating aggregated agreement as a proxy for interpretive validity.

\subsubsection{Bias From Sampling Synthetic Outputs}
A further, often overlooked, source of bias arises from generation procedures—specifically, how researchers sample and select outputs from LLMs. Most models produce responses stochastically, influenced by parameters such as temperature, top-k sampling, and prompt randomness. Yet many studies rely on a single generation per prompt or report only the most “plausible” completion, introducing subjective selection bias into the analysis \cite{atari_which_2023, lee_can_2024}. This is analogous to interviewer effects or coder selection in traditional survey research: choices made during the data generation process can systematically privilege certain response styles over others, potentially narrowing the range of simulated public opinion. To mitigate this, researchers should adopt practices like multi-response sampling, reporting variability across generations, and documenting selection criteria. These procedural safeguards are critical for ensuring that synthetic datasets reflect not just coherent averages but the underlying diversity—and unpredictability—of real human publics.

\subsubsection{Bias from Interaction Effects and Unknown Sources}
Finally, perhaps the most underappreciated source of bias arises from interaction effects among the above factors—and from unknown model dynamics. A mildly leading prompt combined with a stereotyped persona and a biased model may compound distortions in hard-to-predict ways. These nonlinear interactions are difficult to detect with face-validity checks alone and require careful experimental design and validation across multiple axes. As with traditional survey research, the presence of non-apparent biases underscores the need for multi-method triangulation and model audit tools.

\subsection{A Three-Tier Bias Framework to Advance LLM-Augmented Survey}
To clarify how these seven sources of bias relate to one another, we propose a three-tier framework that distinguishes between representational, procedural, and interactional biases, as summarized in Table \ref{tab:bias_types}. Representational biases originate from how identities, viewpoints, and social groups are encoded in the LLM's architecture, training data, and simulated personas. These biases shape which voices are faithfully reproduced and which are distorted or excluded. Procedural biases emerge from the mechanics of simulation itself—how prompts are framed, how outputs are generated and selected, and how fidelity is evaluated. These procedural layers often operate invisibly but can significantly influence the outputs researchers deem “valid” or “accurate.” Finally, interactional biases arise when multiple layers of the system—e.g., a biased prompt interacting with a biased model—combine in nonlinear ways, creating distortions that are difficult to predict or isolate. Grouping the seven sources of bias into these three categories offers a structured way to analyze where bias enters the research pipeline and how it may be mitigated through better model design, prompt engineering, and validation practices.

\begin{table}[ht]
    \centering
    \renewcommand{\arraystretch}{1.2}
    \begin{tabularx}{\textwidth}{>{\bfseries}p{3.5cm} >{\raggedright\arraybackslash}p{4.5cm} X}
        \toprule
        Category & Bias Types & Main Concern \\
        \midrule
        Representational & Persona, Model, Training Data & Who is represented? What viewpoints are reproduced? \\
        Procedural & Prompt, Generation, Evaluation & How is the simulation conducted and judged? \\
        Interactional & Compound/Unknown Effects & What new biases emerge when systems interact? \\
        \bottomrule
    \end{tabularx}
    \vspace{0.5em}
    \caption{Three-Tier Bias Framework}
    \label{tab:bias_types}
\end{table}

Disentangling these multiple sources of bias is crucial for responsible use of LLMs in survey research. Just as survey methodologists developed diagnostics for reliability and bias in human data collection \cite{dillman_mail_2000, paulhus_two-component_1984}, researchers working with LLMs must adopt layered evaluation frameworks that account for prompt variability, demographic fidelity, algorithmic differences, and training data provenance. Encouragingly, recent proposals for role-conditioning, direct bias interrogation, and demographic benchmarking suggest promising paths forward \cite{karanjai_synthesizing_2025, shin_ask_2024}. Ultimately, synthetic data may never be bias-free, but it can be better understood, documented, and calibrated—a prerequisite for its ethical and empirical use in understanding public opinion.

\section{LLMs as Catalysts for Experimental Studies}
The integration of LLMs into experimental research in social sciences marks a methodological shift with transformative potential. This section explores three key ways in which LLMs are reshaping experimental design: as generators of stimuli, as simulated human agents for replicating classic studies, and as social entities within large-scale simulations. Together, these applications reveal how generative AI can go beyond augmenting existing research workflows to actively reconfigure the epistemological premises of experimentation.

\subsection{LLMs as Generators of Experimental Stimuli}
LLMs are reshaping experimental research by enabling the creation of high-quality, scalable, and context-sensitive stimuli. In traditional message effects research, experiments often rely on static materials—such as news stories, political advertisements, or vignettes—presented uniformly to all participants. LLMs allow for the generation of stimuli that are more flexible. They can be personalized, interactive, multimodal, emotionally calibrated, and ethically responsive. Recent studies in communication, political science, and social psychology have demonstrated how LLMs can support new types of content and experimental designs that were previously difficult to implement. This section outlines five key advantages of using LLMs for stimulus generation, based on our review of empirical studies summarized in Table \ref{tab:llm_stimuli}.

\subsubsection{From Universal to Personalized Stimuli: A Methodological Shift}
LLMs make it possible to tailor persuasive messages or informational content to individual participant characteristics—an approach that has been difficult to implement in traditional experimental designs. While the importance of personalization has been widely recognized in message effects research, practical constraints have limited its application at scale. LLMs address this challenge by generating content dynamically, based on participants’ psychological traits, sociodemographic characteristics, political orientations, or prior beliefs.

Several recent studies illustrate this shift. Matz et al. \cite{matz_potential_2024} used GPT-3 to generate messages aligned with participants’ Big Five personality traits, leading to higher persuasive impact than generic messages. Hackenburg and Margetts \cite{hackenburg_evaluating_2024} employed LLMs to produce microtargeted political ads tailored to individual voter profiles. Fisher et al. \cite{fisher_biased_2024} and Goldstein et al. \cite{goldstein_how_2024} found that messages framed to match audience ideology were more effective. Costello et al. \cite{costello_durably_2024} demonstrated that real-time adaptation of message content—based on participants’ expressed beliefs—could enhance engagement in interactive settings. Together, these studies show how LLMs support the design of experiments that incorporate treatment heterogeneity from the outset, reducing the need for post hoc subgroup analysis. In another study, Coppolillo et al. \cite{coppolillo_engagement-driven_2024} trained a LLM to generate content that dynamically adapts to the opinion distribution within a social network. Rather than producing generic messages, the model learns to tailor its output to maximize engagement in specific social contexts, demonstrating a scalable approach to context-sensitive stimulus generation.

\subsubsection{From Static Exposure to Interactive Dialogue: Enabling Dyadic Experimental Designs}
LLMs also enable a new category of interactive experimental designs, where participants engage in real-time, multi-turn conversations with AI-generated agents. Unlike static stimuli, which do not allow clarification or reciprocal exchange, LLM-based dialogues can simulate interpersonal dynamics. This opens new possibilities for studying deliberation, persuasion, and resistance as they occur in interaction.

For example, Costello et al. \cite{costello_durably_2024} used ChatGPT to simulate respective conversations aimed at challenging conspiracy beliefs. The model’s adaptive and empathetic responses led to durable reductions in misinformation endorsement—an outcome rarely achieved through passive exposure. Tessler et al. \cite{tessler_ai_2024} found that AI-facilitated discussions helped participants engage with dissenting views and improved group consensus. Argyle et al. \cite{argyle_leveraging_2023} showed that LLM-mediated conversations promoted civility and mutual understanding in political dialogue. Ju and Aral \cite{ju_collaborating_2025} demonstrated that interactive collaboration with AI agents in creative tasks, such as ad development, improved performance and engagement, illustrating how dyadic human-AI interaction can be integrated into both experimental design and applied communication settings. These studies suggest that experimental stimuli can be reimagined as dynamic social interactions, better reflecting the complexity of real-world communication processes.

\newcommand{\cmark}{\ding{51}} 
\newcommand{\xmark}{} 

\begin{landscape}
\begin{table}[h!]
    \centering
    \renewcommand{\arraystretch}{1.2}
    \footnotesize
    \resizebox{\linewidth}{!}{%
    \begin{tabular}{|p{2.8cm}|p{2.8cm}|p{2.5cm}|p{2.8cm}|p{2.5cm}|c|c|c|c|c|}
        \hline
        \textbf{Study} & \textbf{Focus Domain} & \textbf{Model Used} & \textbf{Stimuli Type} & \textbf{Experiment Scale} & \textbf{Pers.} & \textbf{Inter.} & \textbf{Multi.} & \textbf{A/M Frame} & \textbf{Sensitive} \\
        \hline
        Argyle et al. \cite{argyle_leveraging_2023} & Deliberative dialogue & GPT-3& Dialogue responses & N $\approx$ 1,600 & \cmark & \cmark & \xmark & \cmark & \cmark \\
        Costello et al. \cite{costello_durably_2024} & Misinformation and conspiracy beliefs & GPT-4& Conversational interventions & N = 3,541 & \cmark & \cmark & \xmark & \xmark & \cmark \\
        Coppolillo et al. \cite{coppolillo_engagement-driven_2024} & Engagement optimization & Gemma-2B (RL-fine-tuned) & Social media post completions & N = 1,200 posts & \cmark & \xmark & \xmark & \cmark & \xmark \\
        Fisher et al. \cite{fisher_biased_2024} & AI influence on political decisions & GPT-3.5-turbo & Politically biased feedback in dialogue & N = 299 & \cmark & \cmark & \xmark & \cmark & \cmark \\
        Hackenburg \& Margetts \cite{hackenburg_evaluating_2024} & Political microtargeting & GPT-4 & Tailored political messages & N = 8,587 & \cmark & \xmark & \xmark & \cmark & \xmark \\
        Ju \& Aral \cite{ju_collaborating_2025} & Marketing creativity & GPT-4o + DALL·E 3 & Collaborative text+image ads & N = 2,310; 11,138 ads & \cmark & \cmark & \cmark & \xmark & \xmark \\
        Matz et al. \cite{matz_potential_2024} & Personalized persuasion & GPT-3.5, text-davinci-003 & Trait-tailored persuasive ads & N = 1,788 & \cmark & \xmark & \xmark & \cmark & \xmark \\
        Tessler et al. \cite{tessler_ai_2024} & Democratic deliberation & Chinchilla (fine-tuned) & Consensus-seeking statements & N = 5,734 & \cmark & \cmark & \xmark & \xmark & \cmark \\
        \hline
        Reich \& Teeny \cite{reich_does_2025} & AI as social referent & Perplexity (plus others) & Creative tasks (jokes, poems, stories) & N = 6,801 & \xmark & \cmark & \xmark & \cmark & \xmark \\
        \hline
        Davidson \cite{davidson_auditing_2025} & Multimodal hate speech moderation & GPT-4o, Qwen2-VL & Synthetic multimodal posts & N = 1,854 + 60K pairs & \xmark & \xmark & \cmark & \cmark & \cmark \\
        \hline
        Bär et al. \cite{bar_generative_2024} & Counterspeech and hate speech & LLaMA 3 (70B Chat) & Contextualized counterspeech replies & N = 2,664 & \xmark & \xmark & \xmark & \cmark & \cmark \\
        Bai et al. \cite{bai_explicitly_2025} & Implicit bias in LLMs & GPT-4, Claude-3, LLaMA2, Alpaca & Bias-evaluated prompt tasks & 33,600 prompts & \xmark & \xmark & \xmark & \cmark & \cmark \\
        Goldstein et al. \cite{goldstein_how_2024} & AI-generated propaganda & GPT-3& Propaganda narratives & N $\approx$ 8,221& \cmark & \xmark & \xmark & \cmark & \cmark \\
        Li et al. \cite{li_large_2023} & Emotion-enhanced prompts & ChatGPT-3.5, GPT-4, Vicuna, LLaMA 2, BLOOM, Flan-T5 & Emotion-primed prompts & N = 106 + 45 benchmarks & \xmark & \xmark & \xmark & \cmark & \xmark \\
        \hline
        Jungherr \& Rauchfleisch \cite{jungherr_artificial_2025} & AI in deliberation & None (AI described) & Task descriptions about AI roles & N = 1,850 & \xmark & \xmark & \xmark & \xmark & \cmark \\
        Yin et al. \cite{yin_ai_2024} & Perceived empathy and social support & GPT-4 (via Bing Chat) & Supportive responses to emotional disclosures & N $\approx$ 1,200 & \xmark & \xmark & \xmark & \cmark & \cmark \\
        \hline
    \end{tabular}%
    }

    \vspace{0.5em}
    \begin{minipage}{\linewidth}
        \footnotesize
        \textit{Note.} Pers. = Personalization; Inter. = Interactivity; Multi. = Multimodal; A/M Frame = Affective/Moral Framing; Sensitive = Sensitive Contexts.
    \end{minipage}
    \vspace{0.5em}
    \caption{Overview of Experimental Studies with LLMs as Generators of Experimental Stimuli (Grouped by Primary Dimension)}
    \label{tab:llm_stimuli}
\end{table}
\end{landscape}

\subsubsection{From Text-Only to Multimodal Stimuli: Expanding the Experimental Toolkit}
Recent developments in multimodal LLMs have expanded the scope of experimental research by enabling the generation of both textual and visual content, including images, memes, and synthetic media. This is particularly relevant for communication research, where visual cues play a critical role in shaping perception, emotion, and credibility. Multimodal stimulus generation allows researchers to manipulate both content and presentation format in a coordinated and controlled way.

Davidson \cite{davidson_auditing_2025} used GPT-4o and Qwen2 to create synthetic social media posts that combined text and images in order to study responses to multimodal hate speech. While the models reproduced some patterns observed in human moderation, they also exhibited systematic biases, particularly related to identity categories. This example illustrate how multimodal LLMs enable researchers to examine the interaction of verbal and visual signals, replicate platform-native content formats, and analyze how message form and meaning interact in digital media environments. In a related application, Ju and Aral (2025) \cite{ju_collaborating_2025} conducted a large-scale field experiment in which human workers collaborated with AI agents to produce marketing advertisements. The AI agents generated not only text but also images using DALL·E, forming truly multimodal outputs. These human-AI teams significantly outperformed control groups on objective metrics such as click-through rates. The study demonstrates how multimodal AI collaboration can enhance creative outcomes and operational performance, while also offering researchers new ways to evaluate the communicative and persuasive effects of complex media content in ecologically valid settings.

\subsubsection{From Pretesting to Precision: Controlled Affective and Moral Framing}
A fourth advantage of using LLMs in experimental research is the ability to systematically control emotional tone and moral framing. Traditional designs often face challenges in manipulating affective content without sacrificing message clarity or ecological validity. In contrast, LLMs allow researchers to design prompts that reliably elicit specific emotional responses. This enables controlled testing of the effects of fear, empathy, anger, or moral appeals across different conditions.

Li et al. \cite{li_large_2023} introduced EmotionPrompt, showing that emotional framing enhanced both model output and user receptivity. Goldstein et al. \cite{goldstein_how_2024} and Fisher et al. \cite{fisher_biased_2024} found that emotionally charged political messages generated stronger persuasive effects. Reich and Teeny \cite{reich_does_2025} observed that users reported higher confidence in their own creativity when receiving affirming feedback from an AI, suggesting that affective context influences metacognitive outcomes. Bai et al. \cite{bai_explicitly_2025} and Coppolillo et al. \cite{coppolillo_engagement-driven_2024} further demonstrated how moral language subtly shaped audience perceptions and engagement. Together, these studies highlight the value of LLMs for implementing controlled, replicable affective manipulations—advancing research on emotion, moral reasoning, and message effects.

\subsubsection{From Ethical Constraints to Experimental Exploration: Simulating Sensitive Contexts}
Finally, LLMs allow researchers to study ethically sensitive or operationally challenging scenarios by generating simulated content that avoids direct harm to participants. In areas such as hate speech, disinformation, and online harassment, traditional experiments are often limited by concerns about participant well-being or broader societal risks. LLMs offer a way to construct and control such scenarios, enabling researchers to investigate mechanisms and test interventions without exposing individuals to real-world harm. 

Several recent studies illustrate this potential, while also highlighting critical limitations. Bär et al. \cite{bar_generative_2024} generated both hate speech and counterspeech to test perceptions of offensiveness and message effectiveness, with attention to identity-based framing. They carefully ensured each counterspeech conveyed an appropriate tone, avoided biases, and was culturally sensitive. Goldstein et al. \cite{goldstein_how_2024} simulated authoritarian propaganda in a controlled setting, carefully crafting culturally sensitive language to explore its effects while avoiding real-world harm.

However, caution is warranted when using LLMs to simulate sensitive content. Bai et al. \cite{bai_explicitly_2025} find that LLMs still show significant stereotype biases when evaluated using two psychology-inspired measures, calling for the development of more robust value-aligned benchmarks. Other concerns relate to the reception of AI-generated content. Jungherr and Rauchfleisch \cite{jungherr_artificial_2025} found that AI-authored contribution to deliberative discussion were penalized by participants. Similarly, Yin et al. \cite{yin_ai_2024} found that AI-generated expressions of empathy increased perceived understanding, while this effect weakened when participants were informed that the messages were artificially generated.

These five strengths point to a broader transformation in experimental design. LLMs shift stimuli from static, uniform treatments to dynamic, adaptive, and context-sensitive interventions. They support the development of personalized, dialogic, and multimodal experiments; allow precise control over emotional and moral framing; and enable the study of socially sensitive content under ethical safeguards. Across the studies reviewed, LLMs have increased both the realism and flexibility of experimental methods while opening new possibilities for theory testing and design innovation. For communication researchers, this marks the emergence of a new experimental paradigm—one that is more responsive to the complexity of mediated human behavior in the digital era.

\subsection{Replicating Social Science Experiments with LLM-Simulated Agents}
An emerging trend in experimental research involves using LLMs to simulate human participants and replicate established social science experiments. Instead of recruiting human subjects, researchers prompt models such as GPT-4 or Claude 3.5 to adopt specific demographic profiles and cognitive roles. These simulated participants are then exposed to experimental conditions that mirror prior studies. This approach allows for scalable, low-cost, and highly controlled replications. It enables researchers to test the robustness of findings, examine generalizability, and explore the mechanisms underlying human behavior. We reviewed six LLM-based replication of social science experiments with different focus domains and shared methodological features, as summarized in Table \ref{tab:llm_replication}.

Hewitt et al. \cite{hewitt_predicting_2024} evaluated GPT-4 on 70 preregistered U.S.-based survey experiments from political science, sociology, and communication. The model predicted 476 treatment effects with high accuracy, showing a correlation of 0.85 with observed human results—even for studies not publicly available during LLM's training. However, subgroup-level accuracy declined among underrepresented populations, highlighting ongoing concerns about bias in LLM training data.

Mei et al. \cite{mei_turing_2024} focused on behavioral game experiments. In a series of Turing-style tests, GPT-4 produced decision patterns in games such as the Dictator Game and Ultimatum Game that were statistically indistinguishable from those of human participants. The model also exhibited stable personality traits and consistent decision heuristics, suggesting that it may encode structured social reasoning—at least within well-defined tasks.

Aher et al. \cite{aher_using_2023} introduced a broader framework, termed “Turing Experiments” (TEs), in which LLMs simulate representative populations to test whether canonical findings replicate in silico. Their study reproduced effects from four well-known experiments across psychology, linguistics, and economics—including the Milgram obedience study and garden path sentence interpretation. While results aligned with original findings, the authors identified a “hyper-accuracy distortion,” where the model produced overly idealized responses, raising concerns about ecological validity.

Yeykelis et al. \cite{yeykelis_using_2025} conducted one of the largest replication efforts to date, simulating over 19,000 Claude 3.5-based “AI personas” to assess 133 experimental effects drawn from 45 published studies in media psychology and communication. The models successfully replicated 76\% of main effects and 68\% of all effects. However, replication rates for interaction effects dropped to 27\%, reflecting broader challenges in behavioral science and suggesting limitations in LLMs’ capacity to model complex conditional relationships.

Chen et al. \cite{chen_predicting_2025} extended this approach to field experiments. Using GPT-4 and Claude 3, they tested whether LLMs could predict outcomes from 319 economics field studies involving over 1,600 treatment conclusions. Prompts included abstracts, interventions, and demographic context. Overall accuracy was strong, but performance varied across studies involving gendered, cultural, or identity-based interventions. These results suggest that LLMs may struggle with context-specific reasoning, especially in settings underrepresented in their training data.

Rio-Chanona et al. \cite{rio-chanona_can_2025} present a compelling replication of laboratory market experiments originally designed to study human economic behavior under positive and negative feedback conditions. By translating these experiments into an LLM-simulated environment, they assess whether GPT-3.5 and GPT-4 agents can reproduce the market dynamics observed in human subjects. Their findings show that LLM agents exhibit bounded rationality, a key feature of human behavior, particularly in how they form and update price expectations. 

 All these studies use structured prompting to simulate participants, benchmark model responses against human data, and assess both fidelity and distortion. They vary in prompt complexity, sample size, and domain specificity. Some, like Hewitt et al. \cite{hewitt_predicting_2024} and Aher et al. \cite{aher_using_2023}, focus on lab-based designs; others, such as Chen et al. \cite{chen_predicting_2025}, move into applied, field-based contexts. Together, these studies reveal the flexibility of LLMs as tools for experimental replication, while also underscoring the methodological boundaries that remain—particularly in modeling cultural nuance, intersectionality, and higher-order interactions.

\begin{table}[ht]
    \centering
    \renewcommand{\arraystretch}{1.3}
    \footnotesize
    \resizebox{\linewidth}{!}{%
    \begin{tabular}{|p{3.2cm}|p{3.5cm}|p{2.5cm}|p{3.8cm}|p{3.8cm}|p{2.5cm}|}
        \hline
        \textbf{Study} & \textbf{Focus Domain} & \textbf{Models Used} & \textbf{Replication Scope} & \textbf{Limitations Identified} & \textbf{Type of Experiment} \\
        \hline
        Aher et al. \cite{aher_using_2023} & Psychology, Economics, Social Behavior & GPT-3, GPT-4 & 4 classic experiments (obedience, fairness, language, estimation) & Hyper-accuracy distortion; over-idealized behavior & Lab Experiment \\
        \hline
        Hewitt et al. \cite{hewitt_predicting_2024} & Political Science, Sociology, Survey Experiments & GPT-4 & 70 survey experiments with 476 treatment effects & Lower accuracy for underrepresented groups & Survey Experiment \\
        \hline
        Mei et al. \cite{mei_turing_2024} & Behavioral Economics & GPT-4 & 4 decision-making games (Dictator, Ultimatum, etc.) & Contextual calibration issues in nuanced decision tasks & Lab Experiment \\
        \hline
        Yeykelis et al. \cite{yeykelis_using_2025} & Media Effects, Psychology & Claude 3.5 & 133 effects from 45 studies using 19,000+ personas & Weak replication of interaction effects & Lab \& Field Experiments \\
        \hline
        Chen et al. \cite{chen_predicting_2025} & Field Experiments in Economics & GPT-4, Claude 3 & 319 field experiments with 1,638 conclusions & Prediction skewness across gender, ethnicity, norms & Field Experiment \\
        \hline
        Rio-Chanona et al. \cite{rio-chanona_can_2025} & Experimental Economics, Market Dynamics & GPT-3.5, GPT-4 & 2 core types of human market experiments with positive and negative feedback loops & Limited individual-level heterogeneity; sensitivity to memory length and temperature & Lab Experiment \\
        \hline
    \end{tabular}%
    }
    \vspace{0.5em}
    \caption{Overview of LLM Replication Studies in Experimental Research}
    \label{tab:llm_replication}
\end{table}

These findings support a promising but qualified conclusion: LLMs are capable of modeling many human decision-making patterns and treatment effects, but their reliability varies across domains, demographic contexts, and task complexity. While replications in controlled settings often capture main effects, challenges remain in modeling interaction effects \cite{yeykelis_using_2025}, culturally grounded reasoning \cite{chen_predicting_2025}, and individual variability \cite{aher_using_2023}. 

Rather than serving as a replacement for human experimentation, LLM-based simulation should be understood as a complementary approach to theory testing. It is particularly useful for piloting hypotheses, exploring counterfactuals, and evaluating designs across diverse or underrepresented populations. As this area of research develops, best practices in prompt documentation, demographic representation, and bias auditing will be critical to ensuring both scientific rigor and ethical responsibility.

\subsection{From Replication to Simulation: Expanding the Horizons of Social Modeling}

As the use of LLMs expands in social science research, a growing distinction has emerged between their application in replicating established findings and in simulating novel or unobservable social processes. Replication studies aim to assess whether LLMs can reproduce known experimental effects. In contrast, simulation studies use LLMs to model new social phenomena—often without benchmark data or human baselines. These simulations leverage the generative capacity of LLMs to create agents, model interaction dynamics, and observe collective behavior in contexts that may be too large, complex, or ethically sensitive to study directly. This shift reflects a broader methodological turn: positioning LLMs not only as tools for validation but also as platforms for theoretical exploration, hypothesis generation, and conceptual modeling. 

We reviewed eight simulation studies. While these examples do not capture the full range of empirical work in this area, they offer a useful window into the current state of the field. Table~\ref{tab:llm_sim} provides a detailed comparison of these studies, focusing on aspects such as model selection, simulation structure, memory and interaction mechanisms, and evaluation strategies.

Recent developments in multi-agent simulations using LLMs reflect significant methodological innovation along three key dimensions: model architecture, simulation focus, and environmental complexity. The majority of studies to date have employed proprietary models such as GPT-3.5-turbo and GPT-4, capitalizing on their superior capacity for instruction following and multi-turn reasoning, particularly in domains requiring structured interaction and theory-grounded logic. For instance, simulations involving political negotiation, strategic statecraft, or social contract formation often rely on GPT-4 due to its ability to sustain coherent goal-directed behavior. At the same time, open-source models such as LLaMA 3 and LLaMA3-8B-instruct have begun to gain traction, particularly in studies emphasizing large-scale simulations or modular design. Studies like Yang et al. \cite{yang_oasis_2024} demonstrate that open-source LLMs, when paired with scalable infrastructure, can simulate populations of over one million agents with tractable performance. Several projects benchmark across both proprietary and open-source models to assess performance variation and generalizability.

\begin{landscape}
\begin{table}[ht]
    \centering
    \renewcommand{\arraystretch}{1.2}
    \footnotesize
    \resizebox{\linewidth}{!}{%
    \begin{tabular}{|p{1.3cm}|p{1.6cm}|p{2.5cm}|p{2cm}|p{2.3cm}|p{1.4cm}|p{2.3cm}|p{2.3cm}|p{2.3cm}|}
        \hline
        \textbf{Study} & \textbf{LLM Used} & \textbf{Nature of Simulation} & \textbf{Scale} & \textbf{Focus} & \textbf{Memory} & \textbf{Interaction Type} & \textbf{Environment Complexity} & \textbf{Evaluation / Role Prompting} \\
        \hline
        Chuang et al. \cite{chuang_simulating_2024} & GPT-3.5 & Opinion dynamics in networked agents & Dozens to hundreds & Polarization, confirmation bias, consensus building & Limited & Networked opinion exchange & Networked graph structure & Comparison to models / Political stance prompts \\
        \hline
        Dai et al. \cite{dai_artificial_2024} & GPT-4 & Survival-based sandbox society & Multiple agents over time & Hobbesian theory, sovereign emergence & Yes (multi-day planning) & Survival-based decision-making & Sandbox society & Theory matching / Adaptive planning \\
        \hline
        Guan et al. \cite{guan_richelieu_2024} & GPT-4, LLaMA 3 & Diplomacy game negotiation & 7 agents, multi-round & Strategic negotiation, trust, adaptation & Yes (memory strategy) & Strategic negotiation & Diplomacy board setup & Coherence and trust / National roles \\
        \hline
        Hua et al. \cite{hua_war_2024} & GPT-4 & Historical conflict simulation & 10+ agents & War/peace causality, what-if modeling & Yes (historical trajectories) & War and diplomacy negotiation & Geopolitical setup & Plausibility comparison / Country roles \\
        \hline
        Jin et al. \cite{jin_what_2024} & GPT-4 & Alien civilizations and moral conflict & Variable modular civs & Moral diversity, asymmetric conflict & Yes (civilization modules) & Narrative-based dialogue & Custom modules & Emergent analysis / Alien frameworks \\
        \hline
        Moghimifar et al. \cite{moghimifar_modelling_2024} & GPT-3.5, GPT-4 & Coalition negotiation (MDP) & 2–4 agents & Multilingual, multiparty negotiation & Yes (party history) & Political negotiation with voting & Simulated negotiation & Stability and fairness / Party platforms \\
        \hline
        Park et al. \cite{park_generative_2023} & GPT-3.5 & Sandbox town life simulation & 25 agents & Emergent behavior, memory, planning & Yes (persistent memory) & Daily social interaction & Town with time-space modeling & Behavior analysis / Freeform agents \\
        \hline
        Yang et al. \cite{yang_oasis_2024} & LLaMA3-8B & Social media dynamics (X, Reddit) & 1 million agents & Info diffusion, polarization, herd behavior & Yes (temporal memory) & 21 social media actions & Multi-platform RecSys + time engine & Comparison to real data / Real + generated users \\
        \hline
    \end{tabular}%
    }

    \vspace{0.5em} 

    \begin{minipage}{\linewidth}
        \scriptsize
        \textit{Note.} MDP = Markov Decision Process; RecSys = Recommender System. This table summarizes recent multi-agent LLM simulation studies across domains and levels of complexity.
    \end{minipage}

    \vspace{0.5em} 

    \caption{Overview of LLMs-based Simulation Studies}
    \label{tab:llm_sim}
\end{table}
\end{landscape}

In terms of substantive focus, current research in this space diverges between micro-level and macro-level simulations. Micro-level simulations \cite{chuang_simulating_2024, park_generative_2023} seek to capture emergent behavior grounded in individual-level cognition, memory, and planning. These studies often model daily routines, opinion dynamics, or social coordination, and rely on mechanisms such as memory streams, reflective abstraction, and recursive scheduling. In contrast, macro-level simulations \cite{guan_richelieu_2024, hua_war_2024, moghimifar_modelling_2024} prioritize strategic, institutional, or geopolitical interaction among agents with explicit roles, objectives, and interdependencies. These scenarios require the modeling of trust, alliance formation, deception, and negotiation cycles over multiple rounds. A third category includes theory-driven simulations \cite{dai_artificial_2024, jin_what_2024}, which leverage LLM agents as instruments for exploring normative political theory or ethical divergence, illustrating the utility of such simulations as computational analogs to thought experiments.

The complexity of the simulated environments also varies considerably across studies. Some adopt highly structured settings, such as board games or coalition negotiations, where agents operate under fixed rules and constrained vocabularies. These simulations benefit from well-defined outcome metrics, such as win rates, agreement stability, or classification accuracy. Others implement open-ended or semi-structured environments, including sandbox societies, virtual towns, and multi-platform social networks. These environments necessitate more sophisticated infrastructure—such as time engines, recommender systems, and modular inferencers—to manage dynamic interactions, memory accumulation, and emergent behaviors at scale.

Several methodological trends are evident across this literature. First, memory design has become increasingly central to agent modeling, with innovations including rolling histories, reflective abstractions, and dual-level memory architectures. Second, role prompting is frequently combined with mechanisms for self-reflection, critique, or reinforcement to enhance agent consistency and adaptive behavior. Third, the use of LLM agents to operationalize theoretical constructs—ranging from Hobbesian sovereignty to cross-civilizational ethics—signals the growing importance of simulation as a mode of normative inquiry. Finally, as simulations scale to thousands or millions of agents, engineering advances in modular architecture and computational efficiency are becoming essential. 

Alongside these empirical studies, recent survey papers provide conceptual and methodological overviews of this emerging field. Gao et al. \cite{gao_large_2024}, Mou et al. \cite{mou_individual_2024}, and Wang et al. \cite{wang_survey_2024} offer frameworks for classifying simulation purposes, agent architectures, and evaluation criteria. All three emphasize the importance of distinguishing between simulation layers: individual cognition, interaction-level dynamics, and macro-level outcomes such as polarization or norm formation. Mou et al. \cite{mou_individual_2024} argue that simulations can support both theory building and policy exploration, particularly in settings where direct experimentation is infeasible. Gao et al. \cite{gao_large_2024} broaden the field’s scope to include socio-technical and cyber-physical systems, noting that LLMs are uniquely suited for simulating text-based interaction. Wang et al. \cite{wang_survey_2024} propose a taxonomy of agent design choices—including memory structures, feedback loops, role conditioning, and context awareness—and highlight risks such as response bias, alignment artifacts, and overfitting to prompt patterns.

A shared concern across these reviews is the need for design transparency and standardization. The authors call for clear reporting of prompt formats, scenario construction, and agent logic to support reproducibility and interpretability. They advocate for simulations as middle-range theoretical tools: capable of revealing qualitative dynamics and boundary conditions, but not necessarily suited for exact forecasting.

The studies reviewed—from small-scale simulations of interpersonal coordination \cite{dai_artificial_2024, park_generative_2023} to large-scale modeling of platform dynamics \cite{yang_oasis_2024}—illustrate how LLM-powered simulations are changing the way scholars study collective behavior. These systems do more than mimic human responses; they simulate adaptive reasoning, role interpretation, and symbolic meaning. As Gao, Mou, and Wang emphasize, the key innovation is the use of language as the medium of agent cognition—allowing simulations to incorporate tacit knowledge, identity reasoning, and communicative norms that are often difficult to formalize in traditional models.

Bail \cite{bail_can_2024} similarly argues that generative AI enhances simulation-based research by equipping artificial agents with the capacity to interpret roles, reason with language, and adapt to context. This capability allows researchers to move beyond rule-based modeling and toward simulating how people make sense of complex environments. Across the cases reviewed, this includes agents forming coalitions, negotiating moral disagreement, simulating memory and belief updating, or engaging with platform logics. In domains that are ethically sensitive or operationally challenging—such as misinformation, conflict, or ideological division—LLM-based simulations offer a new approach to exploring counterfactuals, testing theoretical limits, and modeling the unobservable.

For social scientists, these developments point to a growing methodological opportunity. LLM-powered simulations can support the study of not only known patterns, but also hypothetical scenarios, emergent dynamics, and contested spaces of social meaning. As this field evolves, careful attention to agent design, prompt transparency, and evaluative criteria will be critical to ensuring that simulations contribute both theoretical insight and methodological rigor.

\subsection{Recommended Practices for LLM-catalyzed Experiment Designs}
To guide future research, LLMs should be treated not as substitutes for human subjects but as tools that complement human judgment—particularly in the design and interpretation of experiments \cite{demszky_using_2023}. Ensuring rigor and replicability will require the standardization of prompt design, evaluation criteria, and documentation practices. The development of LLM-aware research protocols—such as prompt templates, modified coding schemes, and structured workflows—can help shift current practices from informal experimentation to systematic inquiry.

At the same time, researchers must address emerging risks. These include the homogenization of outputs, cultural overfitting, and reduced variability in open-ended responses \cite{zhang_generative_2025}. It is important to assess not only the face validity of LLM-generated content and behavior, but also potential biases and blind spots—particularly when simulating marginalized identities or sensitive social issues. Transparent reporting of prompts, model versions, and design choices will be critical for building a cumulative and credible body of research.

\section{Re-imagining Lasswell's 5Ws with LLMs}
Harold Lasswell’s classic model—“Who says what, in which channel, to whom, with what effect?”—has long provided a foundational structure for communication research. While often debated and expanded upon, its core dimensions continue to frame inquiries into media content, audience response, and message effects. The emergence of LLMs offers a unique opportunity to revisit each of these components with new methodological tools. This section focuses on three pillars—message, audience, and effect—and discusses how LLMs reconfigure our ability to address Lasswell’s framework with greater flexibility, scale, and theoretical depth.

\subsection{Message Studies with LLMs: From Assumed Agreement to Possible Variations}
In content analysis, the question of \textit{what is being said} has traditionally been answered through manual coding, dictionary-based methods, or supervised classification. These approaches often require simplifying meaning into predefined categories and depend heavily on inter-coder agreement to establish validity. LLMs offer a fundamentally different approach: they can identify topics, frames, sentiment, emotional tone, or even moral values across diverse and large-scale corpora. Recent studies \cite{gilardi_chatgpt_2023, mellon_ais_2023} show that LLMs often match or exceed the performance of human annotators while offering consistency, scalability, and adaptability across tasks.

More importantly, LLMs open the door to modeling interpretive variation. Researchers can now simulate how different readers, annotators, or cultural lenses might interpret the same content, enabling a more intersubjective approach to meaning-making. In our own work \cite{kang_embracing_2025}, we showed how LLMs can be prompted to generate multiple valid interpretations of the same media text, thereby supporting a shift from agreement-based reliability to the study of interpretive diversity. In this way, content analysis becomes not only a method of categorizing meaning but also a tool for exploring how meaning varies across perspectives.

LLMs also make it possible to simulate counterfactual messages—asking, for example, \textit{What if} this article were framed in moral rather than partisan terms? or How would the narrative shift if the author took a different ideological stance? These hypotheticals can now be operationalized and analyzed systematically. LLMs thus allow content analysts to move from documenting existing texts to modeling alternative constructions of discourse, advancing both theory development and hypothesis testing in message studies.

\subsection{Audience Analysis with LLMs: From Distributions to Trajectories}
The second component of Lasswell’s model—\textit{to whom}—has historically been studied through surveys and public opinion research. These methods capture distributions of attitudes and beliefs across population subgroups but are limited by declining response rates, sampling biases, and the difficulty of maintaining long-term longitudinal designs. Digital trace data, while voluminous, tends to offer incomplete views of individuals across fragmented online platforms.

LLMs present a new opportunity: researchers can now simulate individual audience members as dynamic entities, whose beliefs and preferences evolve over time. Synthetic personas can be designed with specific demographic, psychological, or ideological attributes and “exposed” to sequences of messages or interactions. These simulations provide a processual view of how attitudes are formed, challenged, and reshaped—enabling researchers to trace audience trajectories rather than simply categorize audience types.

This logic parallels recent proposals in behavioral science. For instance, Varnum et al. \cite{varnum_large_2024} introduce historical language models trained on text from different time periods to simulate how individuals from past societies might have responded to specific prompts. Although their focus is on cultural reconstruction, the broader insight holds: LLMs can be used to model how cognition and belief unfold over time. In communication research, this offers a way to go beyond static audience segmentation and explore how identity, memory, and exposure interact dynamically.

These affordances also support counterfactual modeling at the audience level. Instead of asking only who currently believes what, researchers can now ask: What would this individual believe had they encountered a different sequence of messages? or How might their opinion have evolved in a more ideologically diverse media environment? Such questions, previously untestable outside of high-cost panel studies, can now be simulated and systematically explored with LLMs.

\subsection{Effect Studies with LLMs: A Road to Counterfactual Reasoning}
The final question in Lasswell’s model—\textit{with what effect}—has long been addressed through experimental designs that test how message features influence outcomes such as attitudes, behaviors, or emotions. These studies have traditionally been limited by the difficulty of creating diverse stimuli, personalizing treatments, and exposing the same individual to multiple conditions. As a result, counterfactual inference—central to causal reasoning—has often remained a theoretical ideal.

LLMs dramatically expand the methodological possibilities for experimental research. They enable researchers to generate stimuli that vary systematically in tone, style, argument structure, or emotional appeal, and to do so across languages, genres, and modalities. Studies have already shown how LLMs can be used to create political messages, health information, interpersonal dialogue, and propaganda with a level of precision and variation not feasible with human-authored stimuli.

More crucially, LLMs make counterfactual experimentation not only feasible but also potentially more valid. Researchers can now simulate how the same synthetic individual would respond to different message conditions—such as varied emotional tone or source credibility—allowing for within-subject comparisons that are rare in real-world designs. This supports a core goal of causal inference: asking \textit{what would have happened if the communicative conditions had been different?}

Still, this potential comes with important limitations. Wu et al. \cite{wu_reasoning_2023} demonstrate that current LLMs often struggle with abstract counterfactual reasoning, performing well on default tasks but significantly worse when the prompt requires imagining hypothetical variants. Likewise, Sen et al. \cite{sen_people_2023} find that while LLMs can generate counterfactual examples of harmful speech, these examples often lack the semantic depth or contrast found in human-authored stimuli. These findings point to the importance of careful validation, prompt design, and researcher oversight when using LLMs for causal experimentation.

Beyond individual-level treatments, LLMs also support multi-agent simulation. Researchers can now test how one message alteration might ripple across a group of simulated agents, enabling the study of emergent dynamics such as deliberation, polarization, or echo chambers. This shifts experimental logic from isolated treatment effects to system-level analysis, expanding the conceptual reach of communication experiments.

Finally, LLMs enable researchers to ask and empirically test a range of \textit{what-if} questions: What if the audience were more politically engaged? What if the message used a moral frame instead of a factual one? These are no longer speculative—they can be operationalized, tested, and iterated across conditions. While limitations remain, especially regarding long-range consistency and reasoning depth, LLMs bring communication effects research closer to its epistemological foundation: using controlled variation to understand how messages shape minds, behaviors, and publics.

\section{Recalibrating the Compass: Integrating LLMs into Methodological Practice}
As generative AI technologies continue to evolve, they present both an opportunity and a challenge for integrating computational tools into social science research. Early work in computational social science primarily focused on analyzing digital traces and large-scale observational data \cite{lazer_computational_2009}. In contrast, the rise of LLMs introduces a new paradigm—one centered not only on observing behavior, but also on simulating, generating, and testing it through interactive designs.

Rather than replacing classical paradigms such as surveys, experiments, and content analysis, LLMs may serve as a bridge between traditional and computational approaches. As Davidson and Karell \cite{davidson_integrating_2025} suggest, generative AI enables new modes of measurement, simulation, and interaction that can augment existing methods without compromising their rigor. Their proposals for concepts such as “interprompt” and “intermodel” reliability, along with efforts to benchmark synthetic data against human-coded ground truth, reflect a broader interest in methodological integration rather than disruption.

Bail \cite{bail_can_2024} further argues that generative AI should be viewed not only as a technical advance but as a conceptual resource. It allows researchers to explore new forms of theorizing by simulating social processes, testing counterfactuals, and modeling interactional dynamics. As reviewed in this paper, LLMs make it possible to generate tailored stimuli, simulate group behavior, and design interactive experiments—advancing a shift from descriptive models to generative experimentation. These developments align with the iterative, abductive reasoning often emphasized in computational social science \cite{davidson_integrating_2025}.

At a more foundational level, Farrell et al. \cite{farrell_large_2025} remind us that LLMs are not just analytic tools—they are cultural and epistemic technologies. Like writing systems or bureaucratic infrastructures, they reconfigure how knowledge is produced, organized, and communicated. This perspective highlights a broader agenda for communication and social science scholars: to study LLMs not only as instruments of research but also as systems that shape how research is done.

In this context, Wing’s \cite{wing_computational_2006} call for “computational thinking” as a core scientific skill becomes increasingly relevant. The key challenge is not simply to adopt LLMs, but to embed them critically into the full research process—from study design and data collection to interpretation and theory building. As the boundaries between human and machine cognition become more porous, communication scholars’ expertise in meaning-making, framing, audience analysis, and discourse will be essential for guiding responsible and reflective use of these technologies.

In sum, the integration of LLMs into social science research is not just about adopting new tools. It requires rethinking how methods are constructed, how theories are tested, and how knowledge is interpreted. This moment calls for a dual commitment: to uphold methodological rigor while embracing the generative potential of AI to expand the scope of empirical and theoretical inquiry.

\bibliographystyle{unsrt}  
\bibliography{references}

\end{document}